\documentclass[10pt,twocolumn,letterpaper]{article}

\usepackage{cvpr}              
\definecolor{cvprblue}{rgb}{0.21,0.49,0.74}
\usepackage[pagebackref,breaklinks,colorlinks,allcolors=cvprblue]{hyperref}
\usepackage{booktabs} 
\usepackage{multirow} 
\usepackage{amsmath}    
\usepackage{breqn}      
\usepackage{xcolor}
\usepackage[accsupp]{axessibility}  
\def\paperID{145} 
\def\confName{CVPR}
\def\confYear{2026}

\title{Redefining Quality Criteria and Distance-Aware Score Modeling for Image Editing Assessment}

\author{
    Xinjie Zhang$^{1}$\thanks{Equal contribution.} \quad 
    Qiang Li$^{1}$\footnotemark[1] \quad 
    Xiaowen Ma$^1$ \quad 
    Axi Niu$^1$ \quad 
    Li Yan$^1$ \quad 
    Qingsen Yan$^{1,2}$\thanks{Corresponding author.} \\
    $^1$Northwestern Polytechnical University, China \\
    $^2$Shenzhen Research Institute of Northwestern Polytechnical University, China \\
    {\tt\small \{zhangxinjie, mozhu87, mxwd, yqs\}@mail.nwpu.edu.cn,} \\
    {\tt\small \{nax, yanli1130\}@nwpu.edu.cn}
}
\begin{document}
\maketitle
\begin{abstract}
Recent advances in image editing have heightened the need for reliable Image Editing Quality Assessment (IEQA). Unlike traditional methods, IEQA requires complex reasoning over multimodal inputs and multi-dimensional assessments. Existing MLLM-based approaches often rely on human heuristic prompting, leading to two key limitations: rigid metric prompting and distance-agnostic score modeling. These issues hinder alignment with implicit human criteria and fail to capture the continuous structure of score spaces. To address this, we propose Define-and-Score Image Editing Quality Assessment (DS-IEQA), a unified framework that jointly learns evaluation criteria and score representations. Specifically, we introduce Feedback-Driven Metric Prompt Optimization (FDMPO) to automatically refine metric definitions via probabilistic feedback. Furthermore, we propose Token-Decoupled Distance Regression Loss (TDRL), which decouples numerical tokens from language modeling to explicitly model score continuity through expected distance minimization. Extensive experiments show our method’s superior performance; it ranks 4th in the 2026 NTIRE X-AIGC Quality Assessment Track 2 without any additional training data.
\end{abstract}    
\section{Introduction}
\label{sec:intro}

Recent advances in image editing models~\cite{Kawar_2023_CVPR,han2024ace,wang2024genartist,liu2025magicquill} have significantly transformed content creation, enabling users to perform complex manipulations with simple instructions. From object insertion and removal to style transfer and attribute modification, modern systems have greatly improved both efficiency and accessibility. However, alongside these advancements arises a critical need for reliable Image Editing Quality Assessment (IEQA), which aims to evaluate whether an edited image faithfully follows the instructions while maintaining visual quality and content consistency.

Compared with traditional Image Quality Assessment (IQA)~\cite{su2020blindly,yang2022maniqa,saha2023re,wang2024large,li2025text}, IEQA introduces fundamentally new challenges. Instead of evaluating a single image in isolation, IEQA requires joint reasoning over multimodal inputs, including the original image, the edited result, and the editing instruction. Moreover, the evaluation is inherently multi-dimensional, typically involving visual quality, editing fidelity, and content preservation~\cite{kawar2023imagic,basu2023editval,li2025balancing}. These characteristics collectively make IEQA a more complex and less explored problem, and current research in this area remains in its early stages.

Recently, several works have explored leveraging Multimodal Large Language Models (MLLMs) to address the IEQA task by exploiting their strong multimodal understanding capabilities to predict quality scores across different dimensions. These methods typically rely on manually designed prompts~\cite{xu2025lmm4edit,ku2024viescore,basu2023editval} and train the models via supervised fine-tuning or reinforcement learning~\cite{cai2025qponderunifiedtrainingpipeline,wu2025visualqualityr1reasoninginducedimagequality,xu2026edithf}. However, they still suffer from two key limitations.

First, \textbf{rigid metric prompting}. In multi-dimensional IEQA, human evaluation is inherently a black-box process~\cite{xu2025largelanguagemodelsactive,ryu2025towards}, where only the final scores are observable while the underlying criteria remain implicit. For different evaluation dimensions, the exact standards followed by human annotators are often unclear. For example, when assessing the visual quality of an edited image, it is ambiguous whether the original image should be considered a reference or how the adherence to instructions should influence the evaluation. Moreover, for MLLMs, even semantically equivalent prompts can lead to significantly different output distributions~\cite{stewart2025surprisinglyfragileassessingaddressing}, further highlighting the instability of such heuristic prompting strategies. However, existing methods usually rely on manually designed prompts~\cite{wu2023q,zhou2022large}, such as ``What is the quality of the edited image?”, which fail to capture and align with the implicit evaluation criteria used by humans.

Second, \textbf{distance-agnostic score modeling}. Directly fine-tuning MLLMs for score prediction using Cross-Entropy (CE) loss treats the task as a classification problem, thereby ignoring the continuous nature of the score space and the intrinsic relationships between different scores~\cite{tang2026revisiting}. Some methods attempt to use semantic words such as ``good" or ``poor" as soft labels~\cite{you2025teaching,wu2023q2}, but this leads to a loss of fine-grained distinctions in the original annotations; for instance, both 3.6 and 3.9 might be mapped to a single ``good" category. Other methods model score distributions over vocabulary tokens~\cite{Tang_Yang_Peng_Wang_Dong_2026,cai2025qponderunifiedtrainingpipeline}, which are not fully compatible with the natural language generation paradigm of MLLMs and require additional post-processing to extract numerical scores from the generated outputs. Although reinforcement learning-based methods can, in principle, model score spaces, they are often computationally expensive and rely on sampling-based rewards~\cite{li2025qinsightunderstandingimagequality,zhao2026reasoningrepresentationrethinkingvisual}, which may lead to discrete or collapsed outputs, thereby limiting score continuity and stability.

To address these issues, we propose Define-and-Score Image Editing Quality Assessment (DS-IEQA). Our key idea is to jointly learn how to define evaluation criteria and how to model score spaces within a multimodal framework. Specifically, we introduce Feedback-Driven Metric Prompt Optimization (FDMPO), which automatically refines metric definitions through probabilistic feedback, enabling the model to better align with implicit human evaluation criteria. In addition, we propose Token-Decoupled Distance Regression Loss (TDRL), which decouples numerical prediction from language modeling and explicitly enforces distance-aware learning in the score space via expected distance minimization.

Extensive experiments demonstrate that our approach achieves more accurate, consistent, and generalizable performance across three evaluation dimensions: visual quality, editing fidelity, and content preservation. Notably, our DS-IEQA ranks 4th in the 2026 NTIRE X-AIGC Quality Assessment Track 2~\cite{ntire26XAIGCqa}, outperforming established baselines without requiring any additional training data.

In summary, our contributions are threefold:
\begin{itemize}
    \item We propose \textbf{DS-IEQA}, a framework that reformulates IEQA as a synergistic task of metric definition learning and continuous score space modeling.
        
    \item We introduce \textbf{FDMPO}, a strategy that automates the refinement of evaluation criteria. By leveraging probabilistic feedback, FDMPO effectively bridges the gap between raw model instructions and implicit human perceptual standards.

    \item We develop \textbf{TDRL}, a loss that explicitly enforces numerical continuity in the score space. This approach mitigates the limitations of standard CE loss, leading to more precise and robust quality assessment.
\end{itemize}

\section{Related Work}
\label{sec:relatedWork}

\subsection{MLLM-based IQA methods}

The emergence of MLLMs has brought about a paradigm shift in image quality assessment (IQA). Unlike traditional IQA methods, which are typically designed to output numerical scores without semantic interpretation, MLLMs demonstrate strong capabilities in understanding high-level visual semantics and generating human-readable feedback. This enables IQA to be formulated as a multimodal reasoning problem, where quality evaluation can be performed in a more interpretable and flexible manner.

To evaluate the perception capabilities of MLLMs for IQA, several benchmark datasets have been proposed, such as Q-Bench~\cite{wu2023q} and AesBench~\cite{huang2024aesbench}. These studies reveal that while MLLMs exhibit promising abilities in visual understanding and coarse-grained quality reasoning, they still struggle with accurate numerical prediction of quality scores. In particular, their outputs often fail to align well with continuous quality annotations, indicating a gap between multimodal reasoning and precise score regression.

This limitation largely stems from a mismatch between the discrete nature of CE loss used in language modeling and the continuous nature of quality scores. To address this issue, Q-Align~\cite{wu2023q2} formulates score prediction as a classification problem over discrete labels, while DeQA-Score~\cite{you2025teaching} further introduces distribution-based soft labels to approximate the Gaussian-like distribution of human scores. However, these methods still rely on predefined representations of score distributions, which introduce implicit assumptions about the relationships between different soft labels and require additional post-processing to recover numerical predictions from distributional outputs. In addition, several methods~\cite{M3-agiqa,xu2025lmm4edit,cai2025qponderunifiedtrainingpipeline} incorporate extra modules to handle the logical outputs of models for score regression; however, they are not fully compatible with the generative nature of MLLMs and rely on additional architectural components.

To overcome this limitation, we propose Token-Decoupled Distance Regression Loss (TDRL) that directly models the continuous structure of the score space without relying on discrete labels,  avoiding the loss of fine-grained numerical information inherent in label-based methods.

\subsection{Image Editing Quality Assessment}

Unlike traditional IQA methods that solely focus on the quality of a single image, image editing quality assessment (IEQA) demands joint reasoning over the original image, the edited image, and the corresponding editing instruction. Furthermore, it involves multi-dimensional evaluation spanning key aspects such as visual quality, editing fidelity, and content preservation~\cite{xu2025lmm4edit,ku2024viescore}.
However, in different evaluation dimensions, the exact standards followed by human annotators are often ambiguous and can be regarded as a black-box process. For instance, when assessing the visual quality of an edited image, it remains unclear whether the original image should serve as a reference. Consequently, clarifying the exact evaluation standards adopted by humans across these dimensions is of great importance.

Existing MLLM-based evaluation methods~\cite{ku2024viescore,zheng2023judgingllmasajudgemtbenchchatbot} typically rely on manually predefined metric prompts. Such rigid metric prompting strategies only learn to generate quality scores at a surface level~\cite{leiter2024prexme,mizrahi2024state}, but fail to capture the implicit and inconsistent criteria inherent in human evaluation~\cite{murugadoss2025evaluating}, which ultimately leads to unstable and untrustworthy scoring results.

To address this challenge, we propose Feedback-Driven Metric Prompt Optimization (FDMPO), which learns to refine metric definitions through probabilistic feedback on score predictions. By optimizing the prompt based on the model’s scoring behavior, FDMPO enables the model to better align with the implicit evaluation criteria used by humans, thereby producing more reliable and consistent quality assessments.

\section{Method}

\subsection{Overview}

\begin{figure*}[ht]
    \centering
    \includegraphics[width=0.965\textwidth]{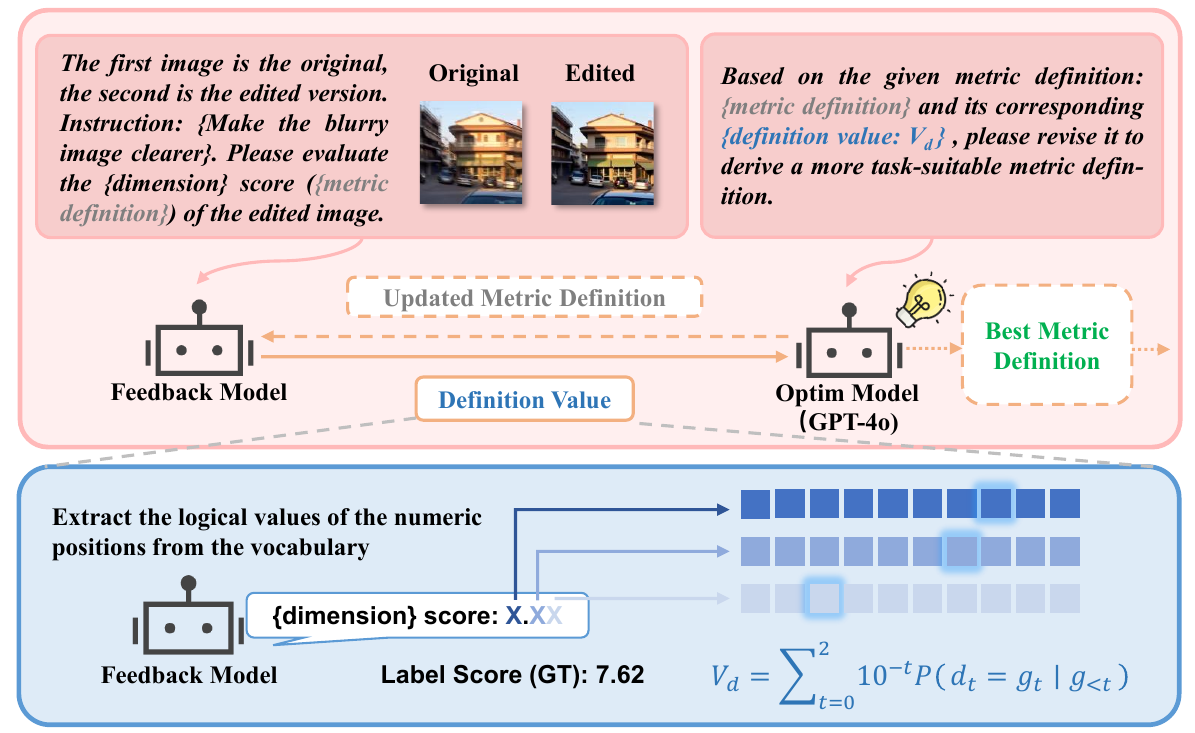}
    \caption{Overview of the proposed Feedback-Driven Metric Prompt Optimization (FDMPO). The framework iteratively refines metric definitions via feedback signals to better align with implicit human evaluation criteria.}
    \label{fig:FDMPO}
\end{figure*}

The IEQA task evaluates an edited image given the original image $x_o$, edited image $x_e$, and editing instruction $ins$. Formally, we learn a function
\begin{equation}
f: (x_o, x_e, ins) \mapsto \mathbf{s} = \{s_{\mathrm{vis}}, s_{\mathrm{edit}}, s_{\mathrm{pres}}\}
\end{equation}
where $\mathbf{s}$ denotes the visual quality, editing fidelity, and content preservation scores.

To address this task, we propose Define-and-Score Image Editing Quality Assessment (DS-IEQA), which jointly learns metric definitions and score prediction. Specifically, we introduce Feedback-Driven Metric Prompt Optimization (FDMPO) to align evaluation with implicit human criteria, and Token-Decoupled Distance Regression Loss (TDRL) to model the continuity of score space without relying on discrete labels. The rest of this section is organized as follows: \Cref{sec:fdmpo} presents FDMPO, \Cref{sec:tdrl} introduces TDRL, and \Cref{sec:training} describes the training pipeline.

\subsection{Feedback-Driven Metric Prompt Optimization (FDMPO)}
\label{sec:fdmpo}

The core motivation for FDMPO comes from the inherent limitations of rigid metric prompting in current image editing evaluators~\cite{wu2023q,zhou2022large}. Human assessment is a complex black-box process~\cite{xu2025largelanguagemodelsactive,ryu2025towards} with implicit criteria, while existing methods rely on manually designed, static prompts that fail to effectively capture these standards. Furthermore, the high sensitivity of MLLMs to prompt variations leads to unstable output distributions~\cite{stewart2025surprisinglyfragileassessingaddressing}, making heuristic instructions unreliable for precise quality assessment. 

To bridge this gap, we transition from manual prompting to FDMPO, which uses a Feedback Model to quantify the quality of a metric definition via its Definition Value ($V_d$), which represents the joint probability of generating the ground-truth numerical tokens. Subsequently, an Optimization Model, specifically GPT-4o, acts as a high-level reasoner to analyze the historical trajectory of $V_d$ and refine the prompt accordingly. This iterative feedback-optimization loop enables the system to converge toward implicit human preferences, effectively aligning the model's scoring logic with human perception.

As shown in \Cref{fig:FDMPO}, initially, an original prompt, comprising evaluation instructions and an input image pair, is fed into the Feedback Model. Instead of treating the output as a simple text string, we perform a fine-grained analysis of the numerical token space. This approach is motivated by the inherent limitations of discrete text generation: standard decoding often results in quantization loss, where multiple images with subtle quality differences are mapped into the same text output; for instance, an MLLM might decode two images with ground-truth scores of 4.4 and 3.6 into the same 4.0 score due to its limited vocabulary. Such discrete outputs fail to provide a sufficiently sensitive gradient or signal for prompt optimization.

To overcome this, we extract the logit values corresponding to each digit position from the model's vocabulary to compute $V_d$. By operating in the logit space, we capture the model's underlying probability distribution, which provides a much denser and more continuous signal than raw text. 

Specifically, the Definition Value $V_d$ is defined as follows:
\begin{equation}
V_d = \sum_{t=0}^{2} 10^{-t} P(d_t = g_t \mid g_{<t})
\label{v_d}
\end{equation}
where the constituent terms are defined as:
\begin{itemize}
\item $t \in \{0, 1, 2\}$ denotes the digit position, representing the integer, tenths, and hundredths places, respectively. The $10^{-t}$ weighting reflects the hierarchical importance of decimal precision, as a correct integer prediction contributes more to alignment confidence than the tenths and hundredths places.
\item $d_t$ and $g_t$ represent the predicted digit and the ground-truth value at position $t$.
\item $g_{<t}$ signifies the ground-truth of preceding digits, such as $g_0$ and $g_1$ in the case of $t=2$.
\item $P(\cdot \mid g_{<t})$ denotes the conditional probability that the model correctly predicts the target digit $g_t$ given the correct preceding context.
\end{itemize}
To ensure robustness and mitigate individual sample noise, $V_d$ is calculated as the mean result aggregated across multiple text-image pairs.

Subsequently, the historical metric definitions and their associated $V_d$ scores are injected into the context of the Optimization Model. Using its in-context learning capabilities, the model compares these trials to find a better optimization path. This process generates an Updated Metric Definition for the next iteration. After multiple iterations, we select the Best Metric Definition by identifying the one that achieved the maximum $V_d$ across all historical trials. This feedback-driven mechanism ensures that the framework progressively identifies evaluation criteria that best align the MLLM's reasoning with human standards.

\begin{figure*}[ht]
    \centering
    \includegraphics[width=1\textwidth]{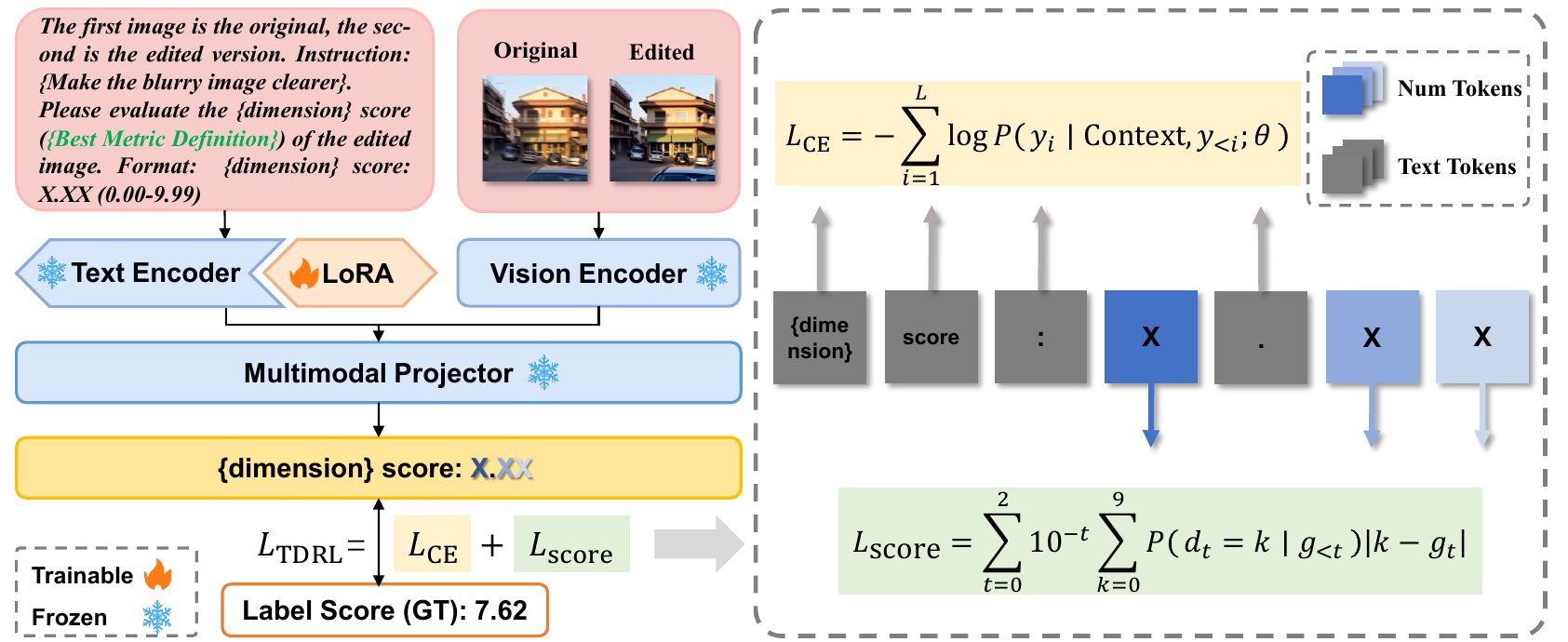}
    \caption{Illustration of Token-Decoupled Distance Regression Loss (TDRL). By decoupling numerical tokens and modeling distance-aware relationships, it enables continuous and precise score prediction without modifying MLLM.}
    \label{fig:TDRL}
\end{figure*}

\subsection{Token-Decoupled Distance Regression Loss (TDRL)}
\label{sec:tdrl}
The motivation for this approach stems from several critical limitations in existing MLLM score prediction methods. Using CE loss treats scoring as a classification task, ignoring the continuous nature of the score space and inherent relationships between scores~\cite{tang2026revisiting}. Semantic soft labels cause the loss of fine-grained differences in original annotations~\cite{you2025teaching,wu2023q2}. Modeling score distributions over vocabulary tokens is not fully compatible with natural language generation and requires extra post-processing. While recent approaches, such as LMM4Edit~\cite{xu2025lmm4edit}, employ an auxiliary score decoder for score-specific fine-tuning, such modifications are incompatible with the LLM’s native causal language modeling, and the reliance on additional modules increases the model's complexity, and fails to leverage the rich semantic knowledge in the LLM's vocabulary. 

To address these issues, we introduce TDRL, which effectively maps numerical regression directly onto the logit space of standard tokens. This ensures full compatibility with the LLM's generative paradigm while eliminating the need for extra structural components. 

As shown in \Cref{fig:TDRL}, we first feed the prompt, whose Best Metric Definition is derived from FDMPO through iterative refinement, and the image pair into the encoder. After encoding them into unified tokens, these representations are processed by a Multimodal Projector, which finally outputs a continuous numerical score. Unlike previous approaches that suffer from distance-agnostic penalties, $L_{\text{TDRL}}$ enables the model to learn to output in the ``{dimension} score: X.XX'' pattern through $L_{\text{CE}}$ and model the distance-aware numerical space to output a score closer to the ground-truth label through $L_{\text{score}}$. The TDRL loss is defined as:
\begin{equation}
	L_{\text{TDRL}} = L_{\text{CE}} + L_{\text{score}}
	\label{l_tdrl}
\end{equation}
For non-numerical tokens, we employ the standard CE loss to ensure linguistic structural consistency:
\begin{equation}
	L_{\text{CE}}=-\sum_{i=1}^{L}\log{P\left(y_i\mid \text{Context},y_{<i};\theta\right)}
	\label{l_ce}
\end{equation}

For the numerical digits, we introduce a distance-aware loss $L_{\text{score}}$ to model the continuous nature of the score space:
\begin{equation}
    L_{\text{score}} = \sum_{t=0}^{2} 10^{-t} \sum_{k=0}^{9} P(d_t=k \mid g_{<t})|k - g_t|\label{l_score}
\end{equation}

Following the hierarchical decomposition in \Cref{v_d}, $t \in \{0, 1, 2\}$ and $10^{-t}$ denote the digit position and its respective magnitude weight, ensuring that the penalty prioritizes integer accuracy over decimal precision. Unlike the alignment-focused $V_d$, $L_{\text{score}}$ evaluates the expected absolute error by traversing all possible digits $k \in \{0, \dots, 9\}$ in the model's vocabulary. Here, $P(\cdot \mid g_{<t})$ represents the predicted probability distribution given the ground-truth prefix $g_{<t}$, while $|k - g_t|$ acts as an ordinal penalty that measures the numerical distance from the ground-truth $g_t$. This design effectively transforms discrete token categories into a continuous, distance-aware score space, ensuring that ``near-miss" predictions are penalized less than distant outliers. By incorporating these positional weights and absolute deviations, the framework naturally aligns with the decimal numeral system and prioritizes the accuracy of higher-magnitude digits.

\subsection{Training Pipeline}
\label{sec:training}

The training pipeline follows two stages to ensure both instruction alignment and numerical precision. Initially, FDMPO is employed to autonomously derive the Best Metric Definition for each of the three evaluation dimensions: visual quality, editing fidelity, and content preservation. These optimized definitions are then integrated into the system prompt in~\Cref{fig:TDRL} to provide a robust semantic foundation for the assessment task. Subsequently, we fine-tune the LLM using Low-Rank Adaptation (LoRA) governed by our proposed $L_{\text{TDRL}}$. This dual-objective loss enables the model to strictly adhere to the prescribed output pattern while generating numerical scores that exhibit high-fidelity alignment with the ground-truth. To reduce interference across dimensions and maintain dimension-specific expertise, we train independent model instances for each of the three evaluation dimensions.
\section{Experiment}

\begin{table*}[t]
  \centering
  \small
  \caption{Comparison on the 2026 NTIRE X-AIGC Quality Assessment Track 2~\cite{ntire26XAIGCqa} and baselines. \textcolor{red}{\textbf{Red Bold}} and \textcolor{blue}{\underline{blue with underline}} indicate the best and second-best performance among participants. Our proposed method achieves competitive performance compared to challenge approaches and baselines using only 8B-scale models without additional training data.}
  \label{tab:participant_metrics}
  \setlength{\tabcolsep}{10pt}
  
  \begin{tabular}{@{}l ccc ccc c@{}}
    \toprule
    \multirow{2}{*}{\textbf{Method}} & \multicolumn{3}{c}{\textbf{In-Distribution}} & \multicolumn{3}{c}{\textbf{Out-of-Distribution}} & \multirow{2}{*}{\textbf{Final}} \\
    \cmidrule(lr){2-4} \cmidrule(lr){5-7} 
    & Visual & Editing & Preserv. & Visual & Editing & Preserv. & \\
    \midrule
    \textit{Baselines} & & & & & & & \\
    LPIPS~\cite{zhang2018unreasonable} & 0.3953 & 0.2199 & 0.5071 & 0.3506 & 0.2508 & 0.5394 & 0.3760 \\
    EditScore~\cite{luo2026editscoreunlockingonlinerl} & 0.3127 & 0.5204 & 0.5560 & 0.2999 & 0.4209 & 0.5875 & 0.4558 \\
    \midrule
    \textit{Challenge Participants} & & & & & & & \\
    helix & 0.7380 & 0.7371 & 0.7563 & 0.6492 & 0.6745 & 0.6935 & 0.7224 \\
    vin1115 & 0.7402 & 0.7372 & 0.7544 & 0.6584 & 0.6778 & 0.6921 & 0.7236 \\
    NTR & \textbf{\textcolor{red}{0.7705}} & 0.7726 & 0.7816 & 0.7100 & 0.7383 & 0.7308 & 0.7603 \\
    UDIQA & \textcolor{blue}{\underline{0.7696}} & 0.7746 & 0.7827 & \textbf{\textcolor{red}{0.7271}} & 0.7344 & 0.7365 & 0.7627 \\
    YuFans & 0.7662 & \textbf{\textcolor{red}{0.7794}} & \textbf{\textcolor{red}{0.7903}} & 0.7092 & \textcolor{blue}{\underline{0.7447}} & \textbf{\textcolor{red}{0.7470}} & \textcolor{blue}{\underline{0.7651}} \\
    \midrule
    \textbf{Ours (DS-IEQA)} & 0.7692 & \textcolor{blue}{\underline{0.7771}} & \textcolor{blue}{\underline{0.7894}} & \textcolor{blue}{\underline{0.7174}} & \textbf{\textcolor{red}{0.7513}} & \textcolor{blue}{\underline{0.7448}} & \textbf{\textcolor{red}{0.7664}} \\
    \bottomrule
  \end{tabular}
\end{table*}
\subsection{Datasets and Evaluation Metrics}
We train our model on the public dataset provided by the 2026 NTIRE X-AIGC Quality Assessment Track 2~\cite{ntire26XAIGCqa}. The model's performance is evaluated independently on three perceptual dimensions: visual quality, editing fidelity, and content preservation.

The performance is measured using both Spearman Rank Correlation Coefficient (SRCC) and Pearson Linear Correlation Coefficient (PLCC) between the predicted scores and the ground-truth MOS. To further assess the robustness and generalization ability of the proposed methods, the validation and test sets are divided into in-distribution and out-of-distribution subsets, where the latter contains edited images generated by image editing models unseen during training. 

For each subset, we compute the average correlation score across three evaluation dimensions by averaging SRCC and PLCC:
\begin{equation}
\begin{aligned}
S_{\mathrm{in}} &= \frac{1}{3}\sum_{d=1}^{3} \frac{\mathrm{SRCC}_d^{\mathrm{in}} + \mathrm{PLCC}_d^{\mathrm{in}}}{2} \\
S_{\mathrm{out}} &= \frac{1}{3}\sum_{d=1}^{3} \frac{\mathrm{SRCC}_d^{\mathrm{out}} + \mathrm{PLCC}_d^{\mathrm{out}}}{2}
\end{aligned}
\end{equation}
The final score is then obtained as a weighted combination of the in-distribution and out-of-distribution performance:
\begin{equation}
\mathrm{Final\ Score} = 0.7\, S_{\mathrm{in}} + 0.3\, S_{\mathrm{out}}
\label{final_score}
\end{equation}

\subsection{Implementation Details}

We adopt Qwen3-VL-8B-Instruct and InternVL3-8B as backbone models and perform parameter-efficient fine-tuning using LoRA. Specifically, LoRA is applied to all layers with a rank of 16 and a dropout rate of 0.05. The models are trained using the AdamW optimizer with a learning rate of $1\times10^{-4}$ and a cosine learning rate schedule with a warmup ratio of 0.03. Training is conducted on two NVIDIA RTX 4090 GPUs with a total batch size of 128.

\subsection{Comparison with Challenge Participants}

We compare our method with participants on the official leaderboard in Table~\ref{tab:participant_metrics}. Our reported results are obtained by averaging the predictions from Qwen and Intern models. Despite using only 8B models and without any additional training data, our method achieves state-of-the-art competitive performance. Moreover, it exhibits consistently balanced performance across both in-distribution and out-of-distribution settings as well as across all three evaluation dimensions, demonstrating strong robustness and generalization capability.

\begin{table*}[t]
  \centering
  \caption{Ablation study of FDMPO and TDRL. Both components contribute to consistent performance gains.}
  \label{tab:ablation}
  \resizebox{\textwidth}{!}{
    \begin{tabular}{@{}c c c c c c c c c@{}}
      \toprule
      \textbf{Model} & \textbf{Setting} & 
      \multicolumn{3}{c}{\textbf{In-Distribution}} & 
      \multicolumn{3}{c}{\textbf{Out-of-Distribution}} & 
      \textbf{Final} \\
      \cmidrule(lr){3-5} \cmidrule(lr){6-8}
      & & Visual & Editing & Preservation & Visual & Editing & Preservation & \\
      \midrule

      Qwen3-VL-8B & w/o FDMPO 
      & 0.724$_{\textcolor{red}{-1.3}}$ & 0.735$_{\textcolor{red}{-1.4}}$ & 0.751$_{\textcolor{red}{-1.6}}$
      & 0.662$_{\textcolor{red}{-1.2}}$ & 0.700$_{\textcolor{red}{-1.4}}$ & 0.701$_{\textcolor{red}{-2.4}}$
      & 0.722 \\
      
      Qwen3-VL-8B & w/o TDRL 
      & 0.681$_{\textcolor{red}{-5.6}}$ & 0.688$_{\textcolor{red}{-6.1}}$ & 0.712$_{\textcolor{red}{-5.5}}$
      & 0.596$_{\textcolor{red}{-7.8}}$ & 0.610$_{\textcolor{red}{-10.4}}$ & 0.642$_{\textcolor{red}{-8.3}}$
      & 0.670 \\

      Qwen3-VL-8B & \textbf{Full Model} 
      & \textbf{0.737} & \textbf{0.749} & \textbf{0.767}
      & \textbf{0.674} & \textbf{0.714} & \textbf{0.725}
      & \textbf{0.737} \\

      \midrule

      InternVL3-8B & w/o FDMPO 
      & 0.732$_{\textcolor{red}{-0.9}}$ & 0.739$_{\textcolor{red}{-1.9}}$ & 0.755$_{\textcolor{red}{-0.9}}$
      & 0.685$_{\textcolor{red}{-2.3}}$ & 0.704$_{\textcolor{red}{-0.8}}$ & 0.714$_{\textcolor{red}{-1.1}}$
      & 0.730 \\
      
      InternVL3-8B & w/o TDRL 
      & 0.686$_{\textcolor{red}{-5.5}}$ & 0.688$_{\textcolor{red}{-7.0}}$ & 0.704$_{\textcolor{red}{-6.0}}$
      & 0.583$_{\textcolor{red}{-12.5}}$ & 0.592$_{\textcolor{red}{-12.0}}$ & 0.606$_{\textcolor{red}{-11.9}}$
      & 0.663 \\

      InternVL3-8B & \textbf{Full Model} 
      & \textbf{0.741} & \textbf{0.758} & \textbf{0.764}
      & \textbf{0.708} & \textbf{0.712} & \textbf{0.725}
      & \textbf{0.743} \\

      \bottomrule
    \end{tabular}
  }
\end{table*}

\subsection{Ablation Study}

\textbf{Impact of FDMPO.} The ablation results in \Cref{tab:ablation} verify the effectiveness of FDMPO in aligning model evaluation with implicit human standards. Removing FDMPO leads to a steady decline across all dimensions. For the Qwen3-VL-8B model, the final average score drops from 0.737 to 0.722. This confirms that manually designed prompts often fail to capture the implicit criteria used by human annotators. Additionally, the decline is particularly evident in the visual and editing dimensions, where human preferences are more subjective and implicit. By iteratively refining metric definitions through a feedback loop, FDMPO effectively bridges the gap between raw model instructions and human-centered evaluation logic.

\begin{figure*}[t]
    \centering
    \includegraphics[width=1\linewidth]{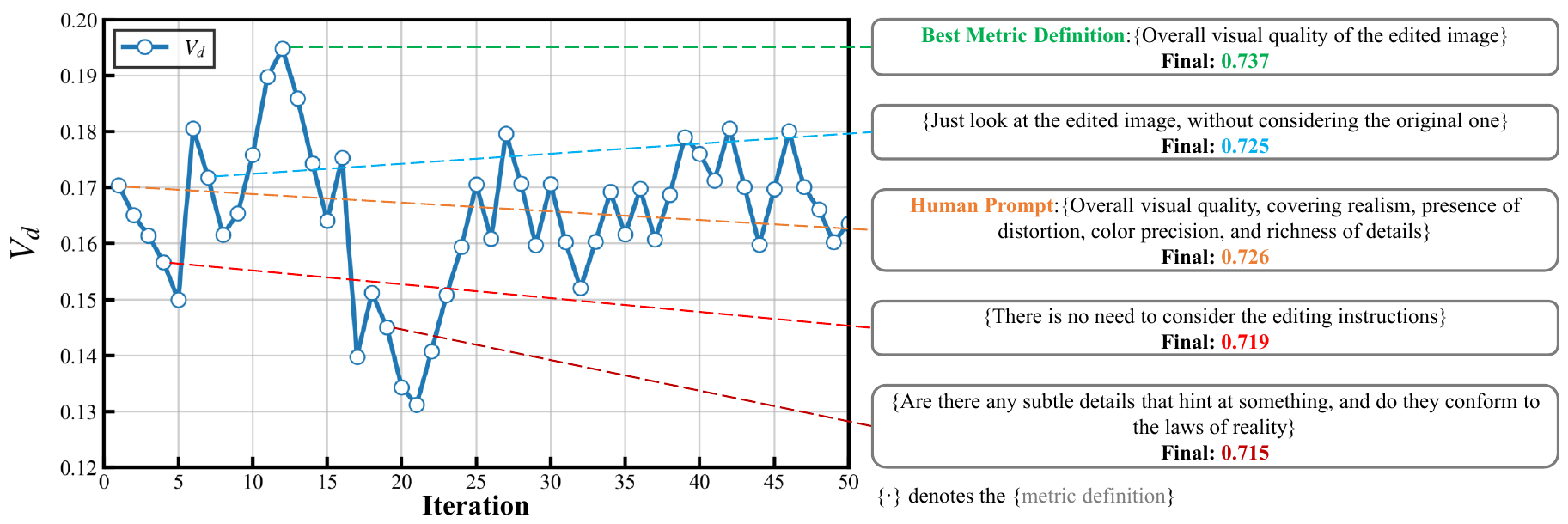}
       \caption{Optimization trajectory of the proposed FDMPO in visual quality. A clear correlation between $V_d$ and final assessment performance is demonstrated. Final Score is calculated following~\Cref{final_score}}
       \label{fig:metric}
\end{figure*}

\noindent\textbf{Validation of $V_d$-Guided Metric Selection.} To evaluate the effectiveness of FDMPO, we visualize the optimization trajectory of the Definition Value ($V_d$) in~\Cref{fig:metric}. The verbal optimization process shows natural fluctuations, as the model occasionally explores less effective reasoning directions. However, by maintaining a history of all trials and selecting the metric definition that maximizes $V_d$, we ensure that the most effective evaluation criteria are preserved. As shown in~\Cref{fig:metric}, the metric definition identified by the peak $V_d$ achieves a superior visual quality score of 0.737, outperforming the original prompt (0.726) proposed in~\cite{xu2025lmm4edit}. We also observe a clear semantic evolution: while lower-scoring prompts often introduce redundant (\eg, ``\textit{overall visual quality, covering realism\dots}") or negative constraints (\eg, ``\textit{no need to consider editing instructions}"), the optimized high-$V_d$ prompts converge toward concise, core perceptual attributes. This semantic distillation explains the model's enhanced generalization; by filtering out linguistic noise and refining the evaluation focus, $V_d$ guides the MLLM to rely on its robust internal visual-textual priors rather than unstable, human-engineered heuristics. The strong alignment between the internal $V_d$ metric and the final assessment scores validates that $V_d$ serves as a reliable proxy, enabling MLLMs to transcend manual heuristics and successfully identify metric definitions that better align with human perceptual reasoning.

\noindent\textbf{Effectiveness of TDRL} The ablation results in \Cref{tab:ablation} confirm that TDRL is the most critical component for numerical precision and model generalization. Removing TDRL leads to a drastic decline in scores across all metrics. For the InternVL3-8B backbone, the final average score drops significantly from 0.743 to 0.663, proving that standard CE loss fails to capture fine-grained numerical relationships. The impact on out-of-distribution scenarios is especially strong without TDRL, with a maximum decrease of 12.5\% in the visual dimension. This suggests that the distance-aware signal enables the model to map latent features to a continuous score space more effectively.

\section{Conclusion}

In this paper, we propose DS-IEQA, which jointly learns metric definitions and score prediction. Specifically, we introduce FDMPO to automatically refine evaluation criteria for better alignment with implicit human judgments, alongside TDRL to explicitly model the continuity of the score space without changing the model architecture. Extensive experiments demonstrate that our method achieves competitive across different evaluation dimensions and data distributions. Notably, our approach achieves state-of-the-art  performance on the 2026 NTIRE X-AIGC Track 2 benchmark~\cite{ntire26XAIGCqa}, demonstrating superior effectiveness and generalizability by utilizing only 8B-scale models without any additional training data. We hope this work provides a simple and effective paradigm for reliable multi-dimensional image editing quality assessment with MLLMs.

\section*{Acknowledgments}
This work was supported in part by the National Natural Science Foundation of China (NSFC) under Grant 62301432 and Grant 6230624, the Natural Science Basic Research Program of Shaanxi under Grant QCYRCXM-2023-057, the Fundamental Research Funds for the Central Universities, and the Guangdong Basic and Applied Basic Research Foundation under Grant 2025A1515011119.
\newpage
{
    \small
    \bibliographystyle{ieeenat_fullname}
    \bibliography{main}
}

\end{document}